\documentclass[conference]{IEEEtran}

\usepackage{cite}

\ifCLASSINFOpdf
  \usepackage[pdftex]{graphicx}
  \graphicspath{{./figs/},{./figs/classic5/},{./}}
\else
  \usepackage[dvips]{graphicx}
  \graphicspath{{./figs/},{./figs/classic5/},{./}}
  \DeclareGraphicsExtensions{.eps}
\fi
\newcommand{\figDir}[1]{./#1} 

\usepackage{amsmath}
\usepackage[mathscr]{euscript} 

\usepackage{algorithmic}

\usepackage{array}

\usepackage[dvipsnames]{xcolor}
\usepackage[hidelinks]{hyperref}
\usepackage{todonotes}
\usepackage{booktabs}
\usepackage{subcaption}
\usepackage{multirow}
\newcolumntype{H}{>{\setbox0=\hbox\bgroup}c<{\egroup}@{}} 

\usepackage{pgfplots}
\pgfplotsset{compat=newest}

\begin{document}

\title{CAS-CNN: A Deep Convolutional Neural Network for Image Compression Artifact Suppression}
\author{\IEEEauthorblockN{Lukas Cavigelli, Pascal Hager, Luca Benini}
\IEEEauthorblockA{Integrated Systems Laboratory, ETH Zurich, Zurich, Switzerland, Email: 
surname@iis.ee.ethz.ch}}

\maketitle

\begin{abstract}
Lossy image compression algorithms are pervasively used to reduce the size of images transmitted over the web and recorded on data storage media. However, we pay for their high compression rate with visual artifacts degrading the user experience. Deep convolutional neural networks have become a widespread tool to address high-level computer vision tasks very successfully. Recently, they have found their way into the areas of low-level computer vision and image processing to solve regression problems mostly with relatively shallow networks. 

We present a novel 12-layer deep convolutional network for image compression artifact suppression with hierarchical skip connections and a multi-scale loss function. We achieve a boost of up to 1.79\,dB in PSNR over ordinary JPEG and an improvement of up to 0.36\,dB over the best previous ConvNet result. We show that a network trained for a specific quality factor (QF) is resilient to the QF used to compress the input image---a single network trained for QF 60 provides a PSNR gain of more than 1.5\,dB over the wide QF range from 40 to 76. 
\end{abstract}

\IEEEpeerreviewmaketitle

\section{Introduction}
Compression methods can be split into two categories: lossless (e.g. PNG) and lossy (e.g. JPEG) \cite{Wang2002}. While lossless methods provide the best visual experience to the user, lossy methods have an non-invertible compression function but can achieve a much higher compression ratio. 
They often come with a parameter to span the trade-off between file size and quality of the decompressed image. In practical uses, lossy compression schemes are often preferred on consumer devices for their much higher compression rate \cite{Wang2002}. 

Particularly at high compression rates, the differences between the decompressed and the original image become visible with artifacts that are specific of the applied compression scheme. These are not only unpleasant to see, but also have a negative impact on many low-level vision algorithms \cite{Yu2016}. Many compression algorithms rely on tiling the images into blocks, applying a sparsifying transform and re-quantization, followed by a generic loss-less data compression \cite{Chew2008}. 

JPEG has become the most widely accepted standard in lossy image compression \cite{Souders2016}, with many efficient software transcoders publicly available and specialized hardware accelerators deployed in many cameras. Due to its popularity, JPEG-compressed images are also widely found on storage devices containing memories of moments experienced with family and friends, capturing the content of historic documents, and holding on to evidence in legal investigations. 

Image compression is also used in wireless sensors systems to transfer visual information from sensor nodes to central storage and processing sites. In such systems, the transmitting node is often battery-powered and thus heavily power-constrained \cite{Kerhet2007}. Transmitting data is often the most expensive part in terms of energy, and strong compression can mitigate this by reducing the required transmit energy at the expense of introducing compression artifacts \cite{Chew2008}. Similar challenges are also seen in mobile devices storing data: size and cost constraints limit the amount of memory for data storage, and the energy available on such devices is depleted rapidly when writing to flash memory---so much that it pays off to apply compression before writing to flash memory \cite{Barr2003,Joo2007}. 
On the processing site, these space and energy constraints are absent and much more computational power is available to decompress and possibly post-process the transmitted or stored images \cite{Chew2008}. 

Deep convolutional neural networks (ConvNets) have become an essential tool for computer vision, even exceeding human performance in tasks such as image classification \cite{He2015}, object detection \cite{Ren2015}, and semantic segmentation \cite{Long2015,Cavigelli2015}. In addition, they have also started to gain relevance for regression tasks in low-level image and video processing, computing saliency maps \cite{Zhao2015}, optical flow fields \cite{Kaluarachchi2015} and single-image super-resolution images \cite{Dong2014} with state-of-the-art performance. 

In this work, we present 1) the construction of a new deep convolutional neural network architecture to remove compression artifacts in JPEG compressed image data, 2) a strategy to train this deep network, adaptable to other low-level vision tasks, and 3) extensive evaluations on the LIVE1 dataset, highlighting the properties of our network and showing that this is the current state-of-the-art performance ConvNet for compression artifact suppression (CAS).

\section{Related Work}
Traditional approaches to suppress compression artifacts can be split into three categories. Various types of intelligent edge-aware denoising such as SA-DCT~\cite{Foi2007,Foi2006}, BM3D~\cite{Dabov2007} have been proposed to address this task during the late 2000s. In recent years, dictionary-based sparse recovery algorithms such as DicTV~\cite{Chang2014}, RTF~\cite{Jancsary2012}, S-D2~\cite{Liu2015a}, $D^3$ \cite{Wang2016}, DDCN~\cite{Guo2016} have achieved outstanding results by directly addressing the deficiencies such as ringing and blocking very specific to JPEG. These algorithms explicitly attempt to optimally reverse the effect of DCT-domain quantization using learned dictionaries very specific to the applied compressor and quantization tables. 

This work was inspired by single-image super-resolution ConvNets, which are a special case of compression artifact removal, where the compression is a simple sub-sampling operation. Several networks have shown to be very successful at this task, such as SRCNN~\cite{Dong2014} or DRCN~\cite{Kim2015}. They use different training procedures and approaches for network construction, but both ConvNets are a simple sequence of convolution and point-wise non-linearity layers. 

Recently, two important works have been published, which apply ConvNets for compression artifact suppression: AR-CNN \cite{Dong2015,Yu2016} and the approach presented in \cite{Svoboda2016}. 
The former starts from the architecture presented in SRCNN. In order to overcome convergence problems, they use transfer-learning from the 4-layer network retrained for artifact reduction to a deeper 5-layer network, as well as between networks trained for different JPEG quality factors (QFs) and datasets. 
In \cite{Svoboda2016} a residual structure extends the simple stacking of convolutional, non-linearity and pooling layers, such that the network is only trained to produce an increment compensating for the distortions. Furthermore, skip elements where some feature maps are bypassing one or multiple layers and are then concatenated to the feature maps at a later stage were introduced. Additionally, they do not use a plain MSE loss function but also include an additional term to emphasize edges. 

The networks of both works were trained on the 400 images contained in the BSDS500 train and test sets and evaluated on the remaining 100 images in the validation set. Testing of these networks was then performed on the LIVE1 dataset (29 images) \cite{sheikh2005live} and, in case of AR-CNN, on the 5 test images of \cite{Foi2007} and a self-collected dataset of 40 photographs from \emph{twitter} as well. We will adopt their test datasets, procedures and quality measures. Our choice of the training dataset is discussed in Section~\ref{sec:dataset}.

\section{Methodology} \label{sec:method}
\begin{figure*}
	\centering
	\includegraphics[width=1.0\linewidth]{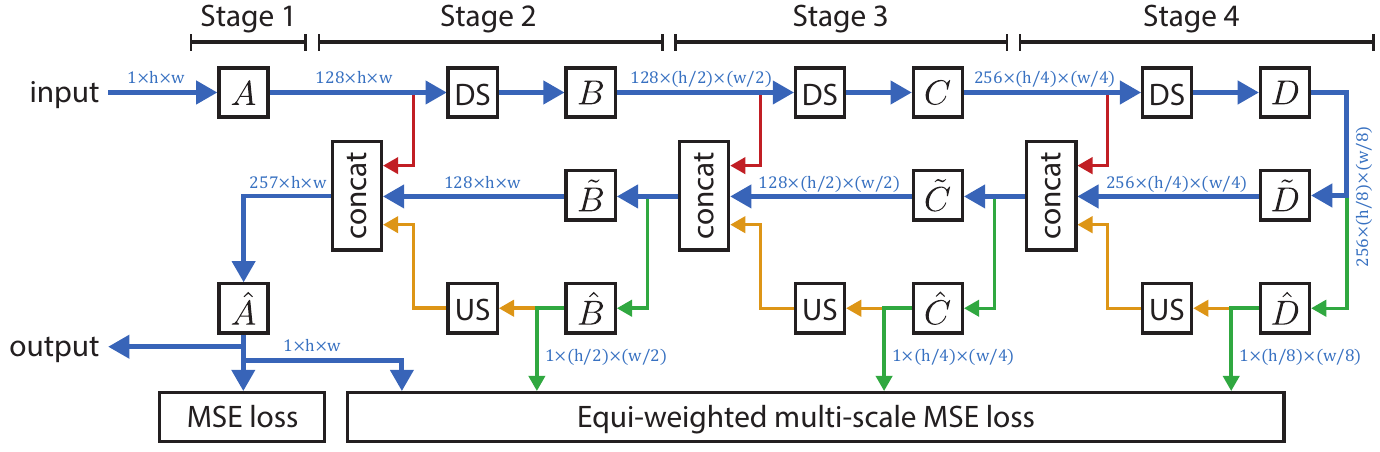}
	\caption{Structure of the proposed ConvNet. The paths are color coded: {\color{NavyBlue}main path (bold)}, {\color{Red}concatenation of lower-level features}, {\color{ForestGreen}multi-scale output paths}, {\color{BurntOrange}re-use of multi-scale outputs}.}
	\label{fig:network}
\end{figure*}
We start from the basic concept of training a deep ConvNet for a regression problem, as has been done for the related task of superresolution~\cite{Dong2014,Kim2015} or other low-level computer vision operations such as optical flow estimation~\cite{Kaluarachchi2015}. The authors of~\cite{Svoboda2016} propose several new elements for artifact reduction ConvNets: A residual architecture, an edge-emphasized loss function, symmetric weight initialization, and skip connections. All these elements were introduced to alleviate the obstacles preventing the training deep networks for regression tasks. 
Taking inspiration from deep neural networks such as FlowNet~\cite{Kaluarachchi2015} and FCN~\cite{Long2015} developed for optical flow estimation and semantic segmentation respectively, we propose a neural network with hierarchical skip connections (cf. Section~\ref{sec:arch}) and a multi-scale loss function (cf. Section~\ref{sec:loss}) for compression artifact suppression. 

\subsection{Network Architecture} \label{sec:arch}
\begin{table}
	\centering
	\caption{Hyperparameters of the Layers}
	\label{tbl:params}
	\begin{tabular}{cccccrHH}
		\toprule
		name & type & \#outp. ch. & \#inp. ch. & filter size & \#param. & \#inp. px. & GOp \\ \midrule
		$A^{(1)}$ & conv & 128 & 1   & $3\times 3$ &   1k & 1049k \\
		$A^{(2)}$ & conv & 128 & 128 & $3\times 3$ & 147k & 1049k \\ 
		$B^{(1)}$ & conv & 128 & 128 & $3\times 3$ & 147k &  262k \\
		$B^{(2)}$ & conv & 128 & 128 & $3\times 3$ & 147k &  262k \\
		$C^{(1)}$ & conv & 128 & 256 & $3\times 3$ & 295k &   65k \\
		$C^{(2)}$ & conv & 256 & 256 & $3\times 3$ & 590k &   65k \\
		$D^{(1)}$ & conv & 256 & 256 & $3\times 3$ & 590k &   16k \\
		$D^{(2)}$ & conv & 256 & 256 & $3\times 3$ & 590k &   16k \\
		$\widetilde D$ & fullconv & 256 & 256 & $4\times 4\,/2$ & 1049k &  16k \\
		$\hat D$  & conv &   1 & 256 & $3\times 3$ &   2k &   16k \\
		$\widetilde C$ & fullconv & 128 & 513 & $4\times 4\,/2$ & 1051k &  65k \\
		$\hat C$  & conv &   1 & 513 & $3\times 3$ &   5k &   65k \\
		$\widetilde B$ & fullconv & 128 & 257 & $4\times 4\,/2$ &  526k & 262k \\
		$\hat B$  & conv &   1 & 257 & $3\times 3$ &   2k &   262k \\
		$\hat A$  & conv &   1 & 257 & $3\times 3$ &   2k &   1049k \\
		\midrule Total &&&&& 5144k \\
		\bottomrule		
	\end{tabular}
\end{table}
An overview of our proposed network is shown in Figure~\ref{fig:network}. The blocks $A,\dots,D$ each consist of two convolutional layers, increasing the number of channels from 1 to 128 and later to 256, the deeper they are in the network. At the same time the resolution is reduced by down-sampling (DS), which is implemented with $2\times 2$ pixel average-pooling layers with $2\times 2$ stride. The main path through the ConvNet (marked blue in Figure~\ref{fig:network}) then proceeds through the full-convolution\footnote{We use the definition of full-convolution (also known as up-convolution, deconvolution, backwards convolution, or fractional-strided convolution) as described in~\cite{Long2015,Noh2015}.} layers $\widetilde{D},\dots,\widetilde{B}$ and the normal convolution layer $\hat{A}$. 
This way we obtain a 12-layer ConvNet, which however cannot be trained to achieve state-of-the-art accuracy using standard training methods. In the following, we list modifications to the network reducing the average path length, allowing it to converge to beyond state-of-the-art accuracy.

To reduce the path length, the higher-resolution intermediate results after each full-convolution layer are enhanced in the next layer by concatenating the lower-level features extracted earlier in the network natively at this resolution (marked red in Figure~\ref{fig:network}). We expect this to benefit the network two-fold: once through the additional information to help infer high-resolution outputs, and second to aid in training the early layers of the network by means of bypassing the middle layers. 

Training deep networks for regression tasks is problematic and while we have reduced the path length for some paths (e.g. $\mathrm{input}\rightarrow A \rightarrow \hat{A}\rightarrow\mathrm{output}$) using the aforementioned method, some very long paths remain. The gradients for adjusting the weights of $D$ are propagated from the output through $\hat{A},\widetilde{B},\widetilde{C},\widetilde{D},D$. 
To improve on this, we introduce a multi-scale optimization criterion: instead of optimizing input-to-output, we reconstruct low-resolution images already from deep within the network using a single convolutional layer (marked green in Figure~\ref{fig:network}), i.e. $\hat{D},\hat{C},\hat{B}$ for 1/64-th, 1/16-th, and 1/4-th of the resolution, respectively. We do not discard the output, but up-sample (US) it by a factor of $2\times$ in each spatial dimension using nearest-neighbor interpolation and concatenate it to the feature maps generated by the full-convolution layer parallel to this path (marked yellow in Figure~\ref{fig:network}). Using this configuration, we have further shortened the deepest stack of layers significantly by reducing the distance from the middle layers to the output.

The parameters of the convolution and full-convolution layers are listed in Table~\ref{tbl:params}. All these layers are followed by a Parametric Rectified Linear Unit (PReLU) \cite{HePReLU2015} activation layer, where the slope for negative inputs is learned from data rather than pre-defined. These units have shown superior performance for ImageNet classification \cite{HePReLU2015}, reducing the issues of dead features \cite{ZeilerVisCNN2014}. 

We have found that learning a residual to the input image instead of the reconstructed image as suggested in previous work \cite{Svoboda2016} did not improve the performance of the proposed ConvNet and thus do not include it in our network. 
The initial weight and bias values are drawn uniformly from the interval $\left(-n_{in}^{-1/2},n_{in}^{-1/2}\right)$, where $n_{in}$ is the number of input channels into that layer.

Batch normalization has shown to reduce the achievable accuracy. The batch-wise normalization of means and variances introduces batch-to-batch jitter thereof into the system, preventing full convergence of the network to the maximum accuracy obtained otherwise.

\subsection{Performance Metrics} \label{sec:metrics}
The most wide-spread performance metrics to evaluate differences between images and many other signals are the mean-squared error (MSE) and the closely related peak signal-to-noise ratio (PSNR). The MSE is the pixel-wise average over the squared difference in intensity between the distorted and the reference image. The PSNR is the MSE normalized to the maximum possible signal values typically expressed in decibel (dB). Following \cite{Dong2015,Svoboda2016} with pixel values normalized to the range $[0,1]$, we use
\begin{align}
	\mathrm{PSNR}(\mathbf{X},\mathbf{\hat{X}}) &= 10 \log_{10}\left(1/\mathrm{MSE(\mathbf{X},\mathbf{\hat{X}})}\right),\\
	\mathrm{MSE}(\mathbf{X},\mathbf{\hat{X}}) &= \left(\sum_{p\in\mathcal{P}}e(\mathbf{x}_p,\mathbf{\hat{x}}_p)^2\right)/\left|\mathcal{P}\right|,
\end{align}
where $\mathcal{P}$ is the set of pixel indexes, $\mathbf{X}$ is the reference image, $\mathbf{\hat{X}}$ is the image to evaluate, and $e$ is the per-pixel error function (e.g. $|x_p-\hat{x}_p|$ for grayscale images). 

Both metrics are fully referenced, comparing individual pixels to the original image and converging to zero for a perfect reconstruction. They are known to differ from perceived visual quality \cite{Wang2004,Sheikh2006,Girod1993,Wang2002} but find wide-spread use due to their simplicity. A variation of the PSNR measure is the IPSNR (increase in PSNR), which is the PSNR difference to the baseline distorted image and thus measures quality improvement. It is also more stable across different datasets. 

A popular alternative is to use the structural similarity index (SSIM) \cite{Wang2004}, which is the mean of the product of three terms assessing similarity in luminance, contrast and structure over multiple localized windows. 
We use the Matlab implementation provided with \cite{Wang2004} for evaluation and use the same parameters as related work \cite{Yu2016,Dong2015,Svoboda2016}: $K_1=0.01, K_2=0.03$, and a $8\times 8$ local statistics window $\mathbf{w}$ of ones.
A third measure used in related work is the PSNR-B~\cite{Yim2011}, which adds a (non-referenced) blocking effect factor (BEF) term to the MSE measure. The BEF measures luminance discontinuities at the horizontally and vertically oriented block boundaries. 
We define the IPSNR-B analogous to the IPSNR. 

\subsection{Loss Function} \label{sec:loss}
During the training of the ConvNets we minimize the MSE criterion, penalizing deviations from the reference image by the squared distance. However, as mentioned in Section~\ref{sec:arch}, in order to improve the training procedure we include not only the full-resolution output, but also the low-resolution outputs from within the network. The reference for these is computed by down-sampling the input image, averaging across 4, 16 and 64 pixels, respectively. Each of these outputs' MSE contributes equally to the overall \emph{multi-scale (MS) loss} function. 

We run the training until convergence with this objective, before removing the lower resolution images from the loss function and continue the training for several epochs to minimize the MSE of only the full-resolution output image (\emph{output loss}), fine-tuning (FT) the network with this optimization objective. 

In previous work, including an edge-emphasized term into the loss function has been proposed \cite{Svoboda2016}. We decided not to introduce such a loss term because it leads to an additional hyperparameter to adjust the weight and because we consider it inconsistent to train the network with a loss function different from the quality measure used to benchmark the results. Tuning the hyperparameters for the best PSNR would result in choosing the weight value of the edge-emphasized loss term to be zero. 

As such, it prevents further improvement in terms of PSNR and SSIM beyond some limit, and the factor with which it is weighted can be used to trade-off overall reconstruction quality and deblocking. We do not include such a term in our setup because our main objective is to maximize the overall reconstruction, which already implies a high-quality deblocking. By training on a large dataset we do not require such a regularization term. 

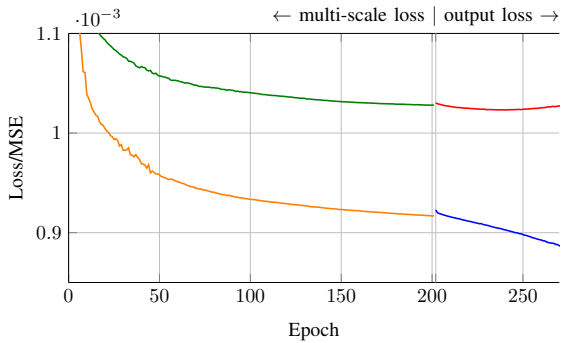
\begin{figure}
	\centering
	\begin{tikzpicture}[scale=0.75]
	\tikzset{mark options={solid, mark size=3, line width=1pt}} %
	\pgfplotsset{
 	 xmin=0, xmax=270
	}
	\begin{axis}[height=6cm, width=1.33\columnwidth-1.5cm, axis y line*=left, xlabel=Epoch, ylabel=Loss/MSE, ymin=0.85e-3, ymax=1.1e-3, grid=major, 
	extra x ticks={202}, extra x tick style={xticklabel pos=right, xticklabels={$\leftarrow$ multi-scale loss $|$ output loss $\rightarrow$\,\,\,\,\,\,\,\,\,\,\,\,}, xmajorgrids=true}] 
		\addplot [thick, color=orange] table [x=id, y=mse_train, col sep=comma] 
			{\figDir{accEvolution.csv}};
		\addplot [thick, color=blue] table [x=id, y=mse_train_ft, col sep=comma] 
			{\figDir{accEvolution.csv}};
		\addplot [thick, color=green!50!black] table [x=id, y=mse_test, col sep=comma]
			{\figDir{accEvolution.csv}};
		\addplot [thick, color=red] table [x=id, y=mse_test_ft, col sep=comma]
			{\figDir{accEvolution.csv}}; 
	\end{axis}
	\end{tikzpicture}
	\caption{Loss improvement by number of training epochs for compression with quality factor 20. It is split horizontally into a phase with the multi-scale loss function and one for fine-tuning with the output loss. The green and red curves are the output loss on the test set. The yellow and blue curve show the loss on the training set. Note that the yellow curve is showing the multi-scale loss and is scaled up by a factor of 3 to fit within the value range of the figure. An epoch during the fine-tuning phase contains 150k instead of 50k images.}
	\label{fig:energyEffThroughput}
\end{figure}

\subsection{Dataset} \label{sec:dataset}
\begin{table}[b!]
	\centering
	\caption{Restoration Quality Comparison on LIVE1}
	\label{tbl:qualityComparison}
	\begin{tabular}{clcccc}
		\toprule
		QF & Algorithm     	& & PSNR\,[dB]	& PSNR-B\,[dB]	& SSIM \\ 
		
		\midrule \multirow{6}{*}{10}
		& JPEG				& \cite{MATLAB2015a} & 27.77		& 25.33 		& 0.791 \\
		& SA-DCT			& \cite{Foi2007} & 28.65		& 28.01 		& 0.809 \\
		& AR-CNN			& \cite{Yu2016} & 29.13		& 28.74 		& 0.823 \\
		& L4				& \cite{Svoboda2016} & 29.08		& 28.71 		& 0.824 \\
		& ours, MS loss		& & 29.36		& 28.92 		& 0.830 \\ 
		& ours, w/ loss FT 	& & \textbf{29.44}& \textbf{29.19}& \textbf{0.833} \\ 
		
		\midrule \multirow{7}{*}{20}
		& JPEG				& \cite{MATLAB2015a} & 30.07			& 27.57 		& 0.868 \\
		& SA-DCT			& \cite{Foi2007} & 30.81			& 29.82 		& 0.878 \\
		& AR-CNN			& \cite{Yu2016} & 31.40			& 30.69 		& 0.890 \\ 
		& L4				& \cite{Svoboda2016} & 31.42			& 30.83 		& 0.890 \\
		& L8				& \cite{Svoboda2016} & 31.51			& \textbf{30.92}& 0.891 \\
		& ours, MS loss		& & 31.67			& 30.84 		& 0.894 \\
		& ours, w/ loss FT	& & \textbf{31.70}& 30.88 		& \textbf{0.895} \\ 
		
		\midrule \multirow{6}{*}{40}
		& JPEG				& \cite{MATLAB2015a} & 32.35		& 29.96 		& 0.917 \\
		& SA-DCT			& \cite{Foi2007}	 & 32.99		& 31.79 		& 0.924 \\
		& AR-CNN			& \cite{Yu2016} & 33.63		& 33.12 		& 0.931 \\
		& L4				& \cite{Svoboda2016} & 33.77		& -- 			& --    \\
		& ours, MS loss 	& & 33.98		& 32.83 		& 0.935 \\ 
		& ours, w/ loss FT 	& & \textbf{34.10}& \textbf{33.68}& \textbf{0.937} \\ 
		
		\midrule \multirow{2}{*}{60}
		& JPEG				& \cite{MATLAB2015a} & 33.99		& 31.89 		& 0.940 \\
		& ours, w/ loss FT 	& & \textbf{35.78}& \textbf{35.10}& \textbf{0.954} \\ 

		\midrule \multirow{2}{*}{80}
		& JPEG				& \cite{MATLAB2015a} & 36.88		& 35.47 		& 0.964 \\
		& ours, w/ loss FT 	& & \textbf{38.55}& \textbf{37.73}& \textbf{0.973} \\ 
		\bottomrule
	\end{tabular}
\end{table}
Previous networks for compression artifact reduction were trained on the 400 train and test images of the BSDS500 dataset and tested on the 100 remaining validation images \cite{Yu2016,Dong2015,Svoboda2016}. The authors of \cite{Svoboda2016} show that this is the limiting factor for further improvement of their larger L8 network with 220k learned parameters. We do not want to constrain the size of our network by the amount of available training data, particularly since we do not need hard-to-obtain labels for it.
We thus use the large, widely-known and publicly available ImageNet~2013 detection dataset~\cite{Deng2009}, which consists of 396k training and 20k validation color images of various sizes. From each image we take cut-outs of $120\times 120$ pixels to generate our dataset. 

The color images are transformed to YCbCr space and only the luminance channel is used further. The input to the network is then generated by compressing the resulting single-channel image using the Matlab JPEG compressor\footnote{We have used this compressor to remain comparable with related work. Other implementations such as libjpeg or libjpeg-turbo use different quantization tables and, in case of these two libraries, result in a significantly larger file size and as a consequence also a better PSNR for the same quality factor.} with a bit depth of 8. 

For training our network we take 50k images of the $120\times 120$ pixel cut-outs from the training set and 10k cut-outs for the validation set. We increase the size of the training set to 150k for fine-tuning with the output loss function. Testing is performed on the 29 images of the LIVE1 dataset. 

We use the Torch framework~\cite{Collobert2011} with cuDNN~v5.1.3~\cite{Chetlur2014} for our evaluations. We optimize the network parameters with Adam \cite{Kingma2015} starting with a learning rate of $10^{-4}$. A minibatch size of 20 images was used and training was parallelized over two Nvidia Titan~X Maxwell GPUs. We have not applied any preprocessing to the images before feeding them into the network.
Our main training was conducted for quality factor 20 compressed input data and we have trained the networks for other quality factors starting from this one to reduce training time. For the forward pass, a throughput of 1.01\,Mpixel/s has been measured with a Nvidia GTX1080 using single-precision floating-point operations.

\section{Results \& Discussion}
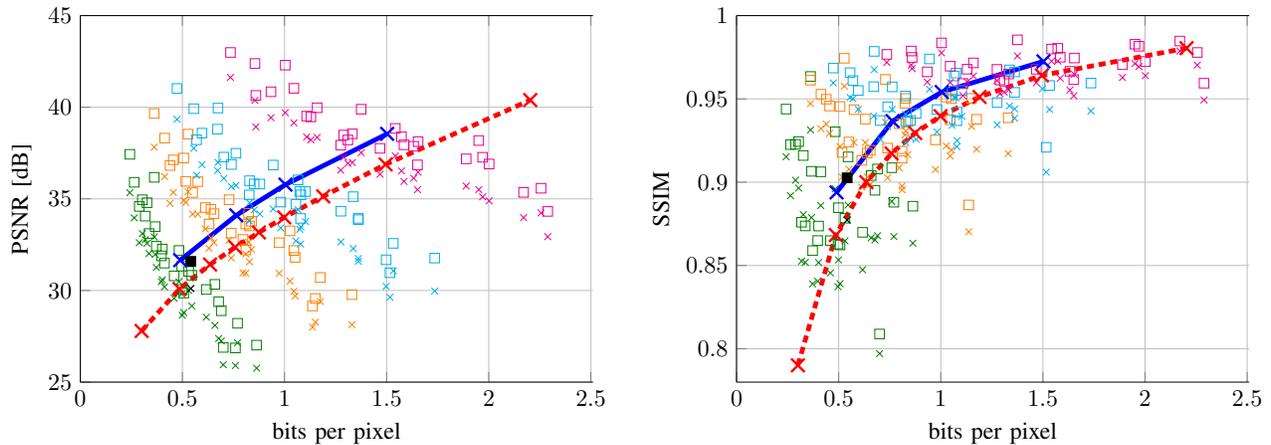
\begin{figure*}[tb]
	\centering
    \begin{subfigure}[b]{0.47\textwidth}
		\begin{tikzpicture}[scale=0.9]
		\tikzset{mark options={solid, mark size=4, line width=1pt}}
		\pgfplotsset{
	 	 xmin=0, xmax=2.51
		}
		\begin{axis}[height=7cm, width=1.25\columnwidth-1.5cm, axis y line*=left, xlabel=bits per pixel, ylabel={PSNR [dB]}, ymin=25, ymax=45, grid=major , every axis legend/.append style={nodes={right}}]
			
			\addplot [only marks, mark=square, mark size=2, mark options={green!50!black}]
				table [x=bitPerPixel, y=psnr, col sep=comma]  
				{\figDir{perImageMetrics_qf20.csv}};
			\label{lgnd:qf20_out}
			\addplot [only marks, mark=x, mark size=2, mark options={green!50!black}]
				table [x=bitPerPixel, y=psnr_in, col sep=comma]  
				{\figDir{perImageMetrics_qf20.csv}};
			\label{lgnd:qf20_in}
			
			\addplot [only marks, mark=square, mark size=2, mark options={orange}]
				table [x=bitPerPixel, y=psnr, col sep=comma]  
				{\figDir{perImageMetrics_qf40.csv}};
			\label{lgnd:qf40_out}
			\addplot [only marks, mark=x, mark size=2, mark options={orange}]
				table [x=bitPerPixel, y=psnr_in, col sep=comma]  
				{\figDir{perImageMetrics_qf40.csv}};
			\label{lgnd:qf40_in}
			
			\addplot [only marks, mark=square, mark size=2, mark options={cyan}]
				table [x=bitPerPixel, y=psnr, col sep=comma]  
				{\figDir{perImageMetrics_qf60.csv}};
			\label{lgnd:qf60_out}
			\addplot [only marks, mark=x, mark size=2, mark options={cyan}]
				table [x=bitPerPixel, y=psnr_in, col sep=comma]  
				{\figDir{perImageMetrics_qf60.csv}};
			\label{lgnd:qf60_in}
			
			\addplot [only marks, mark=square, mark size=2, mark options={magenta}]
				table [x=bitPerPixel, y=psnr, col sep=comma]  
				{\figDir{perImageMetrics_qf80.csv}};
			\label{lgnd:qf80_out}
			\addplot [only marks, mark=x, mark size=2, mark options={magenta}]
				table [x=bitPerPixel, y=psnr_in, col sep=comma]  
				{\figDir{perImageMetrics_qf80.csv}};
			\label{lgnd:qf80_in}
			
			\addplot [only marks, black, mark=square*, mark size=2, mark options={black}] 
				coordinates {(0.54149, 31.586)}; 
			\label{lgnd:lh3}
			\addplot [only marks, black, mark=x, mark size=2, mark options={black}] 
				coordinates {(0.54149, 30.108)}; 
			\label{lgnd:lh3_jpg}
				
			\addplot [thick, densely dashed, line width=2pt, mark=x, mark size=0, color=red] 
				table [x=bitPerPixel, y=psnr, col sep=comma] 
				{\figDir{jpegSweep.csv}};
			\label{lgnd:jpeg}
			\addplot [thick, line width=2pt, mark=x, mark size=0, color=blue] 
				table [x=bitPerPixel, y=psnr, col sep=comma] 
				{\figDir{perImageMetrics_means.csv}};
			\label{lgnd:mean}
		\end{axis}
		\end{tikzpicture}
		\label{subfig:psnrBpp}
    \end{subfigure}%
    \begin{subfigure}[b]{0.47\textwidth}
		\begin{tikzpicture}[scale=0.9]
		\tikzset{mark options={solid, mark size=4, line width=1pt}}
		\pgfplotsset{
	 	 xmin=0, xmax=2.51
		}
		\begin{axis}[height=7cm, width=1.25\columnwidth-1.5cm, axis y line*=left, xlabel=bits per pixel, ylabel={SSIM}, ymin=0.78, ymax=1.0, grid=major , every axis legend/.append style={nodes={right}}] 
		
			\addplot [only marks, mark=square, mark size=2, mark options={green!50!black}]
				table [x=bitPerPixel, y=ssim, col sep=comma]  
				{\figDir{perImageMetrics_qf20.csv}};
			\addplot [only marks, mark=x, mark size=2, mark options={green!50!black}]
				table [x=bitPerPixel, y=ssim_in, col sep=comma]  
				{\figDir{perImageMetrics_qf20.csv}};
			
			\addplot [only marks, mark=square, mark size=2, mark options={orange}]
				table [x=bitPerPixel, y=ssim, col sep=comma]  
				{\figDir{perImageMetrics_qf40.csv}};
			\addplot [only marks, mark=x, mark size=2, mark options={orange}]
				table [x=bitPerPixel, y=ssim_in, col sep=comma]  
				{\figDir{perImageMetrics_qf40.csv}};
			
			\addplot [only marks, mark=square, mark size=2, mark options={cyan}]
				table [x=bitPerPixel, y=ssim, col sep=comma]  
				{\figDir{perImageMetrics_qf60.csv}};
			\addplot [only marks, mark=x, mark size=2, mark options={cyan}]
				table [x=bitPerPixel, y=ssim_in, col sep=comma]  
				{\figDir{perImageMetrics_qf60.csv}};
			
			\addplot [only marks, mark=square, mark size=2, mark options={magenta}]
				table [x=bitPerPixel, y=ssim, col sep=comma]  
				{\figDir{perImageMetrics_qf80.csv}};
			\addplot [only marks, mark=x, mark size=2, mark options={magenta}]
				table [x=bitPerPixel, y=ssim_in, col sep=comma]  
				{\figDir{perImageMetrics_qf80.csv}};
				
			\addplot [only marks, black, mark=square*, mark options={black}] 
				coordinates {(0.54149, 0.90258)};
			\addplot [only marks, black, mark=x, mark options={black}] 
				coordinates {(0.54149, 0.87728)};
				
			\addplot [thick, densely dashed, line width=2pt, mark=x, mark size=0.1, color=red] 
				table [x=bitPerPixel, y=ssim, col sep=comma] 
				{\figDir{jpegSweep.csv}};
			\addplot [thick, line width=2pt, mark=x, mark size=0.1, color=blue] 
				table [x=bitPerPixel, y=ssim, col sep=comma] 
				{\figDir{perImageMetrics_means.csv}};
		\end{axis}
		\end{tikzpicture}
		\label{subfig:ssimBpp}
    \end{subfigure}
	\caption{PSNR (left) and SSIM (right) evaluated on the LIVE1 dataset with respect to the number of bits per pixel required to store the compressed image. The ordinary JPEG performance is shown as (\ref{lgnd:jpeg}) for QF~10 to 90 in steps of 10, averaged over all images in the dataset. Individual images are shown with markers: ordinary JPEG (\ref{lgnd:qf20_in}), after CAS-CNN (\ref{lgnd:qf20_out}). The image depicted in Figure~\ref{fig:visualComp} is marked with (\ref{lgnd:lh3} and \ref{lgnd:lh3_jpg}). The different quality factors are color coded: QF~20 (\ref{lgnd:qf20_out}), QF~40 (\ref{lgnd:qf40_out}), QF~60 (\ref{lgnd:qf60_out}), QF~80 (\ref{lgnd:qf80_out}). The CAS-CNN output quality averaged over the dataset is shown as (\ref{lgnd:mean}).}
	\label{fig:psnrBppPlot}
\end{figure*}

\begin{figure*}
	\centering
	\def\subfigImpWidth{0.45\textwidth}
    \begin{subfigure}[b]{\subfigImpWidth}
		\centering
		\begin{tikzpicture}[scale=0.75]
		\tikzset{mark options={solid, mark size=3, line width=1pt}} %
		\pgfplotsset{
	 	 xmin=0, xmax=100.1
		}
		\begin{axis}[height=6cm, width=1.33\columnwidth-1.5cm, axis y line*=left, xlabel=quality factor, ylabel={IPSNR [dB]}, ymin=0, ymax=2, grid=major, 
		extra x ticks={202}] 
			\addplot [thick, smooth, mark=x, green!50!black] table [x=qfIn, y=ipsnr_qfNet20, col sep=comma]{\figDir{outp-ipsnrEval.csv}};
			\label{plt:ipsnrQf20}
			\addplot [thick, smooth, mark=o, orange] table [x=qfIn, y=ipsnr_qfNet40, col sep=comma]{\figDir{outp-ipsnrEval.csv}};
			\label{plt:ipsnrQf40}
			\addplot [thick, smooth, mark=square, cyan] table [x=qfIn, y=ipsnr_qfNet60, col sep=comma]{\figDir{outp-ipsnrEval.csv}};
			\label{plt:ipsnrQf60}
			\addplot [thick, smooth, mark=triangle, magenta] table [x=qfIn, y=ipsnr_qfNet80, col sep=comma]{\figDir{outp-ipsnrEval.csv}};
			\label{plt:ipsnrQf80}
		\end{axis}
		\end{tikzpicture}
	\end{subfigure}
    \begin{subfigure}[b]{\subfigImpWidth}
		\centering
		\begin{tikzpicture}[scale=0.75]
		\tikzset{mark options={solid, mark size=3, line width=1pt}} %
		\pgfplotsset{
	 	 xmin=0, xmax=100.1
		}
		\begin{axis}[height=6cm, width=1.33\columnwidth-1.5cm, axis y line*=left, xlabel=quality factor, ylabel={IPSNR-B [dB]}, ymin=0, ymax=4, grid=major, 
		extra x ticks={202}] 
			\addplot [thick, smooth, mark=x, green!50!black] table [x=qfIn, y=ipsnrb_qfNet20, col sep=comma]{\figDir{outp-ipsnrEval.csv}};
			\addplot [thick, smooth, mark=o, orange] table [x=qfIn, y=ipsnrb_qfNet40, col sep=comma]{\figDir{outp-ipsnrEval.csv}};
			\addplot [thick, smooth, mark=square, cyan] table [x=qfIn, y=ipsnrb_qfNet60, col sep=comma]{\figDir{outp-ipsnrEval.csv}};
			\addplot [thick, smooth, mark=triangle, magenta] table [x=qfIn, y=ipsnrb_qfNet80, col sep=comma]{\figDir{outp-ipsnrEval.csv}};
		\end{axis}
		\end{tikzpicture}
	\end{subfigure}
	\caption{PSNR and PSNR-B improvement for various compression quality factors for networks trained with images compressed with a single quality factor: QF~20~(\ref{plt:ipsnrQf20}), QF~40~(\ref{plt:ipsnrQf40}), QF~60~(\ref{plt:ipsnrQf60}), QF~80~(\ref{plt:ipsnrQf80}), evaluated on the LIVE1 dataset.}
	\label{fig:multiQfEval}
\end{figure*}
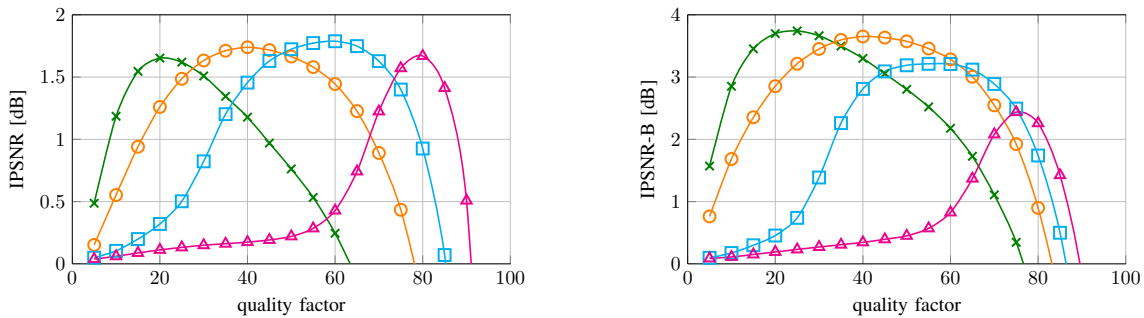

\begin{figure*}[p]
	\centering
	\def\subfigWidth{0.32\linewidth} 
	\def\subfigWidthB{0.33\linewidth}
    \begin{subfigure}[b]{\subfigWidth}
		\includegraphics[width=\linewidth]{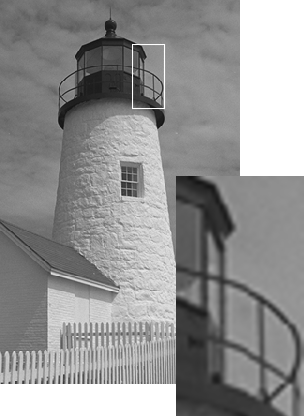}
        \caption{uncompressed}
        \bigskip
        \label{fig:wide-visual-lighthouse3-original}
    \end{subfigure}
    \hspace*{\fill}
    \begin{subfigure}[b]{\subfigWidth}
		\includegraphics[width=\linewidth]{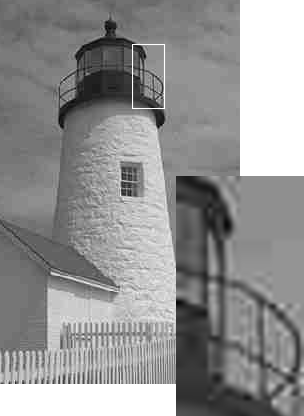}
        \caption{compressed (JPEG~QF\,20)}
        \bigskip
        \label{fig:wide-visual-lighthouse3-distorted}
    \end{subfigure}
    \hspace*{\fill}
    \begin{subfigure}[b]{\subfigWidthB}
		\includegraphics[width=\linewidth]{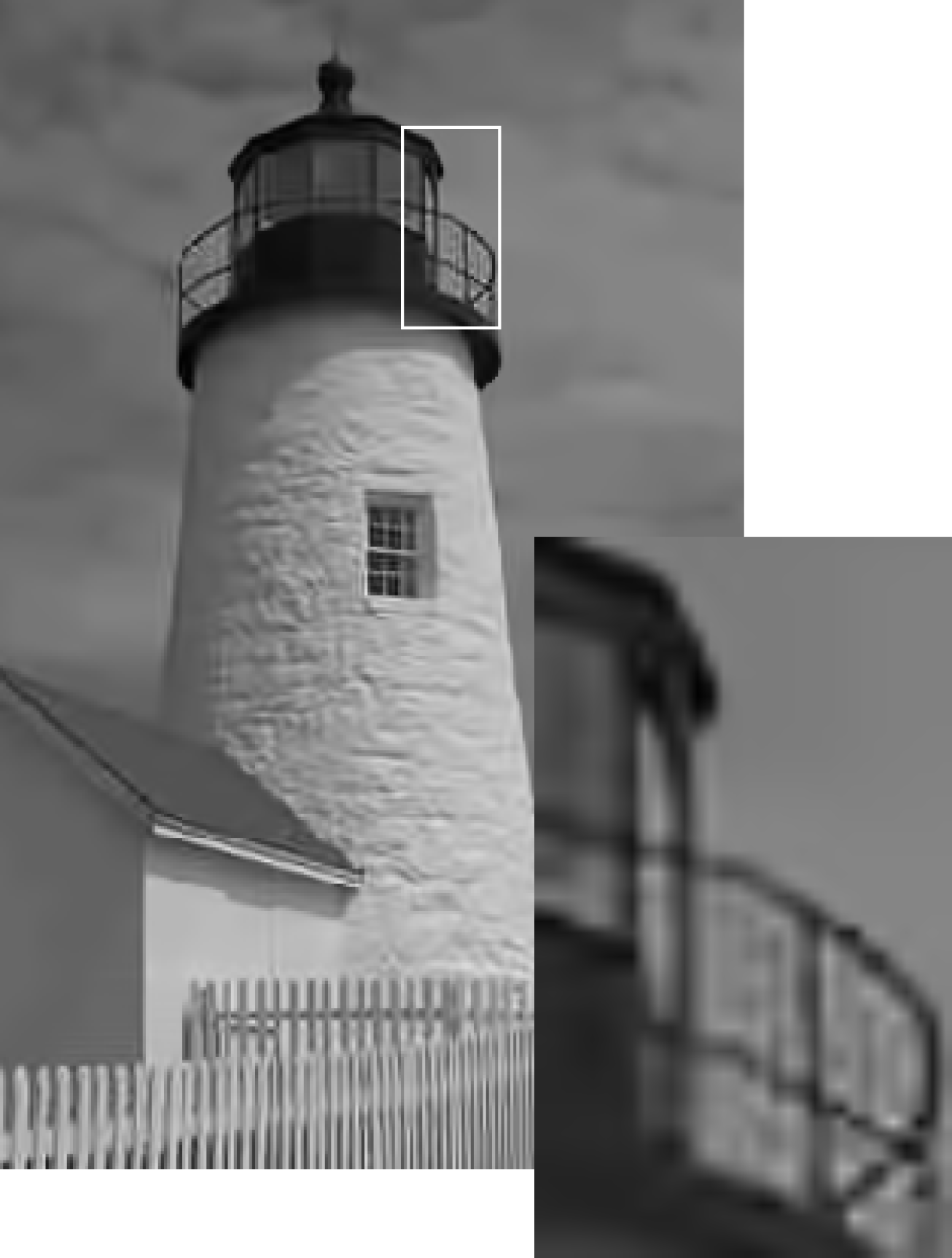}
        \caption{SA-DCT}
        \bigskip
        \label{fig:wide-visual-lighthouse3-sadct}
    \end{subfigure}

    \begin{subfigure}[b]{\subfigWidth}
		\includegraphics[width=\linewidth]{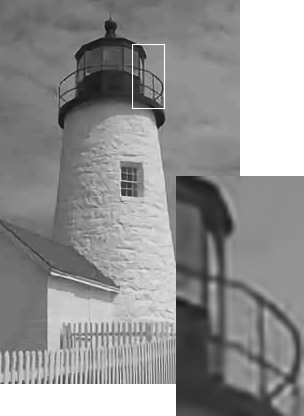}
        \caption{AR-CNN}
        \bigskip
        \label{fig:wide-visual-lighthouse3-arcnn}
    \end{subfigure}
    \hspace*{\fill}
    \begin{subfigure}[b]{\subfigWidth}
		\includegraphics[width=\linewidth]{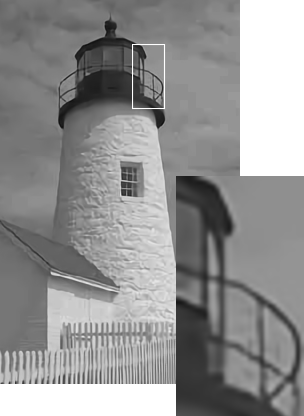}
        \caption{L8}
        \bigskip
        \label{fig:wide-visual-lighthouse3-L08}
    \end{subfigure}
    \hspace*{\fill}
    \begin{subfigure}[b]{\subfigWidthB}
		\includegraphics[width=\linewidth]{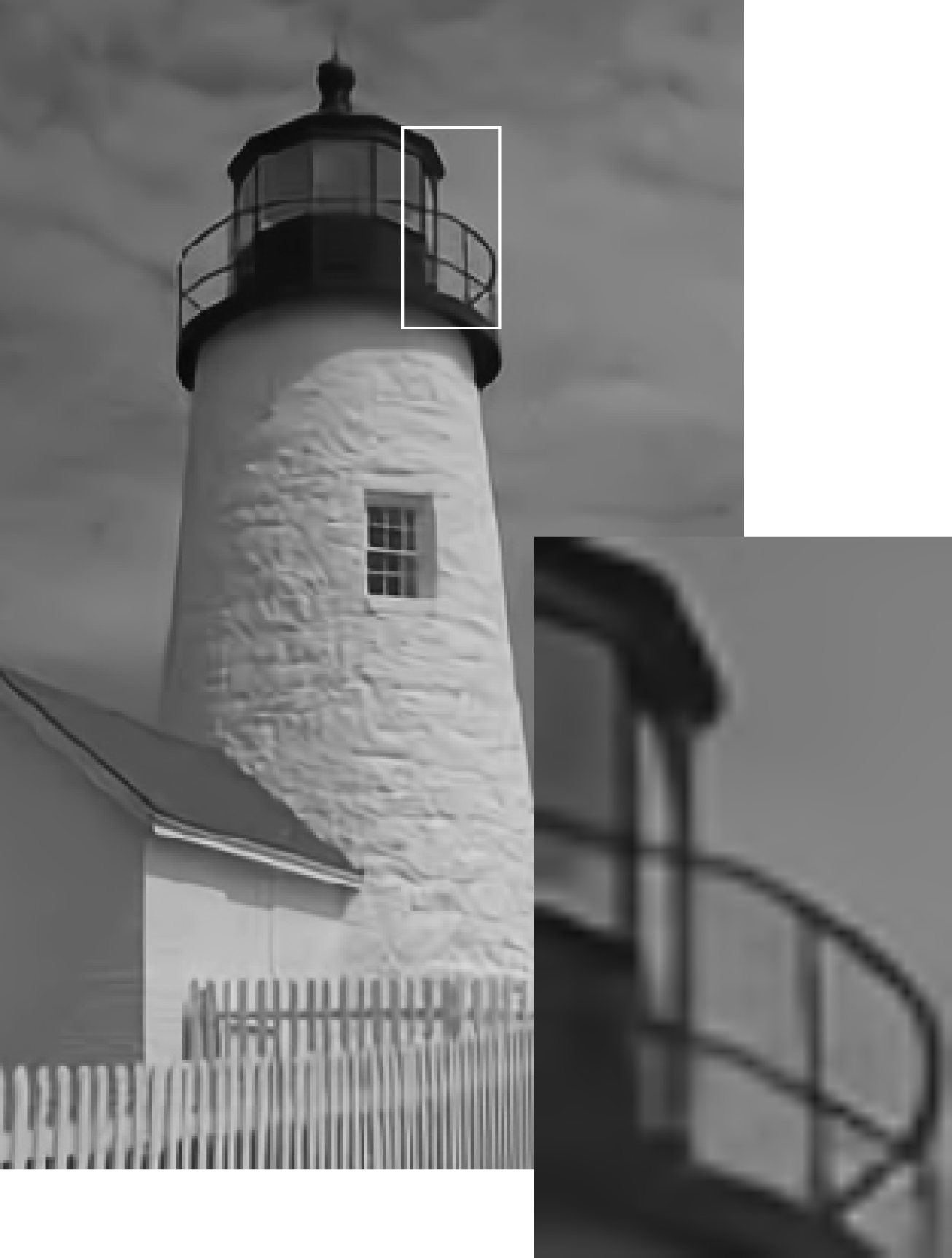}
        \caption{CAS-CNN (ours)}
        \bigskip
        \label{fig:wide-visual-lighthouse3-ours}
    \end{subfigure}
    \caption{Qualitative comparison of reconstruction quality on the \emph{lighthouse3} image of the LIVE1 dataset for JPEG quality factor~20. Images (a),(b),(d),(e) reprinted with permission from \cite{Svoboda2016}.}
    \label{fig:visualComp}
\end{figure*}
We have evaluated the mean PSNR, PSNR-B and SSIM across the LIVE1 dataset for the JPEG quality factors 10, 20, 40, 60 and 80, and compare them to related work in Table~\ref{tbl:qualityComparison}. We use the same JPEG compressor as in AR-CNN~\cite{Dong2015} and Svoboda \emph{et al.}~\cite{Svoboda2016} (i.e. Matlab), with which we obtain the identical baseline PSNR of 30.07\,dB for QF~20 and 27.77\,dB for QF~10 for the JPEG compressed image with respect to the uncompressed reference. 

For our network, we list results directly after training with the multi-scale loss function as well as after fine-tuning with the output loss. The already state-of-the-art results are further improved by this two-step learning procedure. 
Overall, we can see a significant improvement in PSNR of 0.19\,dB over the L8 network \cite{Svoboda2016}, 0.30\,dB over AR-CNN and 1.63\,dB over ordinary JPEG for QF~20. The SSIM is also improved to 0.895. For QF~10 we see a gain of 1.67\,dB over ordinary JPEG, 0.36\,dB over the L4 network and 0.31\,dB over AR-CNN, the state-of-the-art ConvNet for this configuration. 

For QF~10, we improve the PSNR-B by 0.45\,dB over previous work. However, for a lower compression rate, we do not exceed the PSNR-B value achieved by the L8 network. As described in the next paragraph, there are no visible blocking artifacts after applying our ConvNet. PSNR-B has been introduced for benchmarking deblocking algorithms, and by its definition the blocking artifact-penalizing term measuring the differences between pixels along the block boundary does not vanish even for a perfect reconstruction. An image with higher reconstruction quality might thus suffer from a lower PSNR-B value because of clearer edges all over the image including at the block boundaries. 

In Figure~\ref{fig:psnrBppPlot} we show the distribution of the individual images of the LIVE1 dataset in terms of PSNR and SSIM with respect to the used number of bits per pixel for several QFs. The average PSNR and SSIM for each QF is also shown, visualizing that this method works for strong as well as for weak compression. Looking at the individual images, it becomes visible that our method improves not only the mean PSNR and SSIM, but enhances each individual image. 

\begin{figure*}[p]
	\centering
	\def\subfigWidth{0.325\linewidth} 
    \begin{subfigure}[b]{\subfigWidth}
		\includegraphics[width=\linewidth]{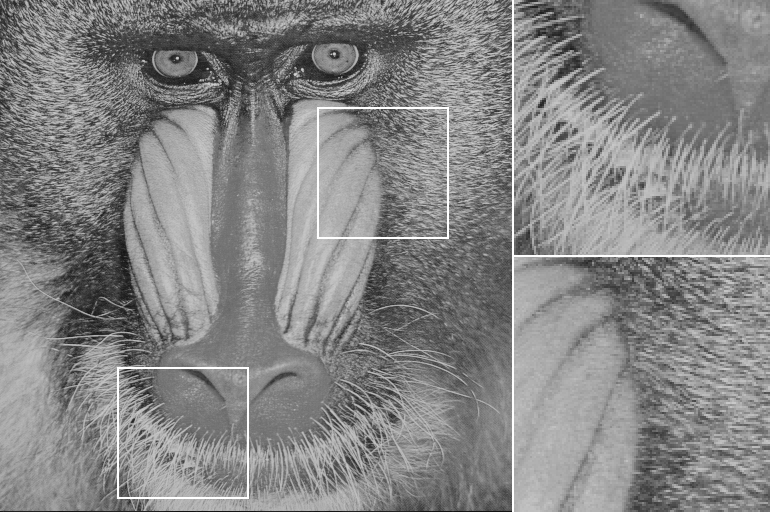}
        \caption{uncompressed}
    \end{subfigure}%
    \hspace*{\fill}
    \begin{subfigure}[b]{\subfigWidth}
		\includegraphics[width=\linewidth]{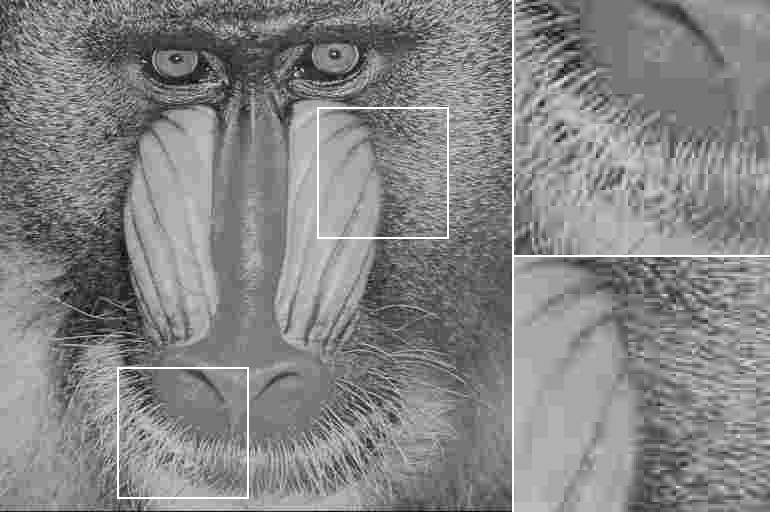}
        \caption{JPEG QF\,10 24.333/22.104/0.7093}
    \end{subfigure}%
    \hspace*{\fill}
    \begin{subfigure}[b]{\subfigWidth}
		\includegraphics[width=\linewidth]{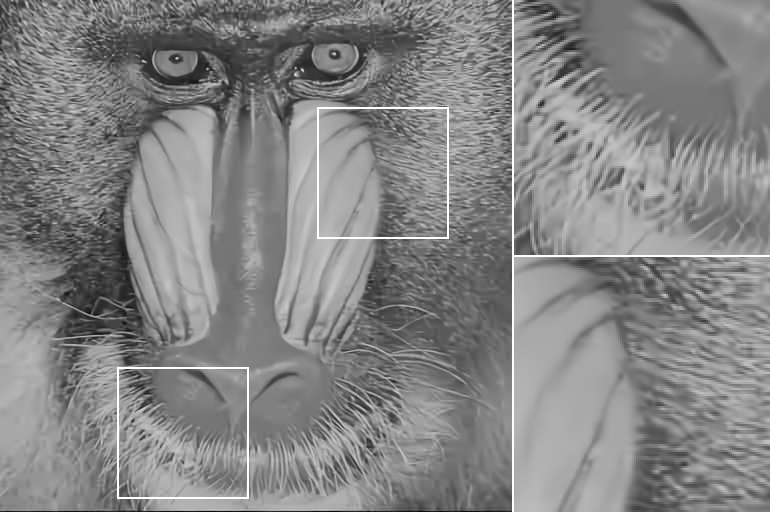}
        \caption{CAS-CNN 25.159/24.746/0.7310}
    \end{subfigure}%
\\
    \begin{subfigure}[b]{\subfigWidth}
		\includegraphics[width=\linewidth]{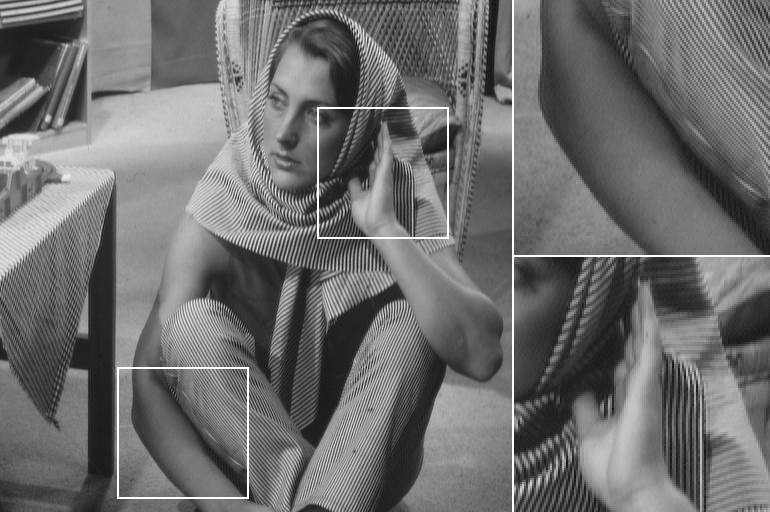}
        \caption{uncompressed}
    \end{subfigure}%
    \hspace*{\fill}
    \begin{subfigure}[b]{\subfigWidth}
		\includegraphics[width=\linewidth]{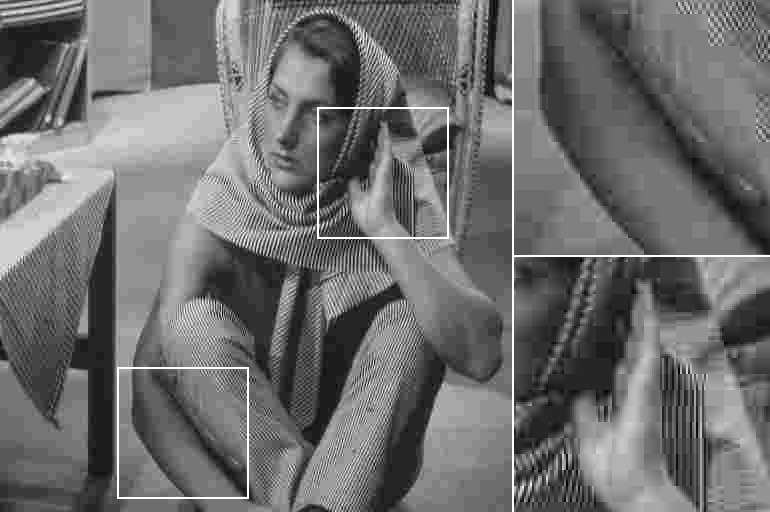}
        \caption{JPEG QF\,10 25.788/23.484/0.7794}
    \end{subfigure}%
    \hspace*{\fill}
    \begin{subfigure}[b]{\subfigWidth}
		\includegraphics[width=\linewidth]{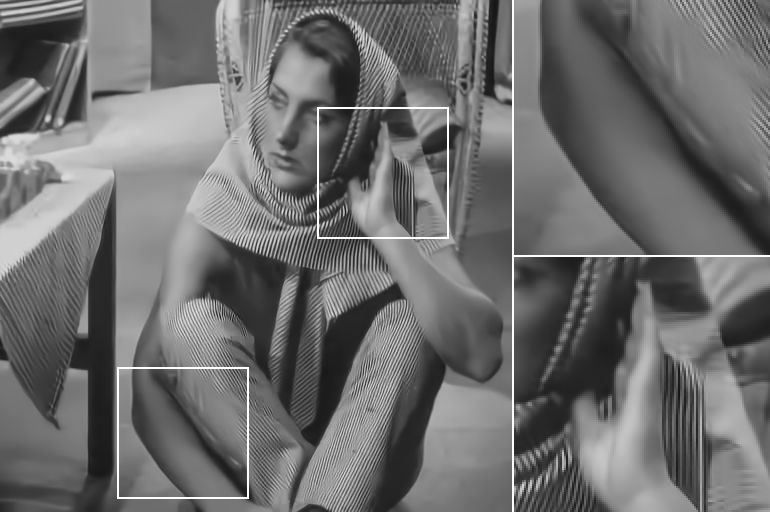}
        \caption{CAS-CNN 28.200/27.612/0.8499}
    \end{subfigure}%
\\
    \begin{subfigure}[b]{\subfigWidth}
		\includegraphics[width=\linewidth]{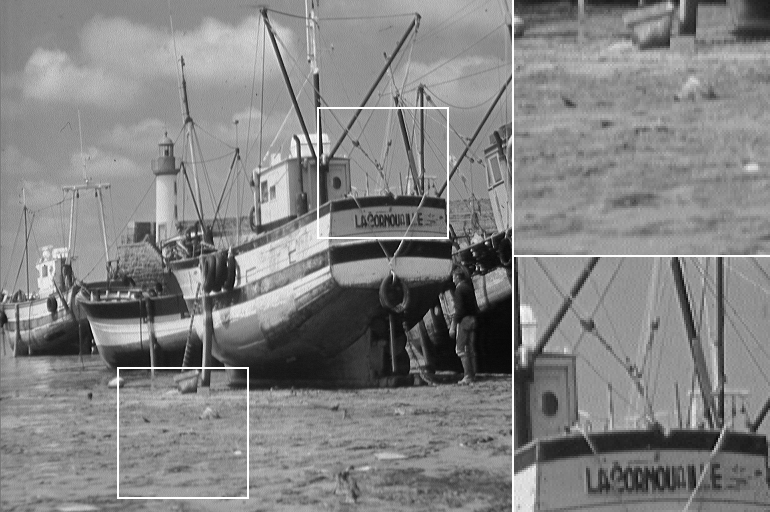}
        \caption{uncompressed}
    \end{subfigure}%
    \hspace*{\fill}
    \begin{subfigure}[b]{\subfigWidth}
		\includegraphics[width=\linewidth]{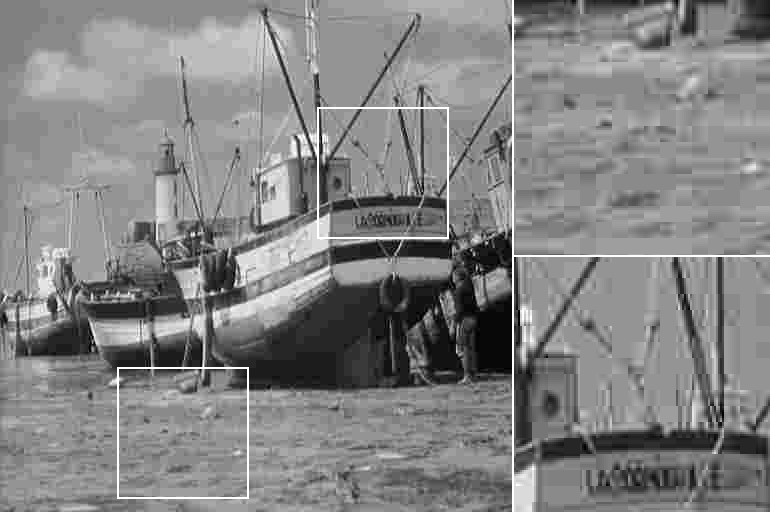}
        \caption{JPEG QF\,10 28.135/25.505/0.7801}
    \end{subfigure}%
    \hspace*{\fill}
    \begin{subfigure}[b]{\subfigWidth}
		\includegraphics[width=\linewidth]{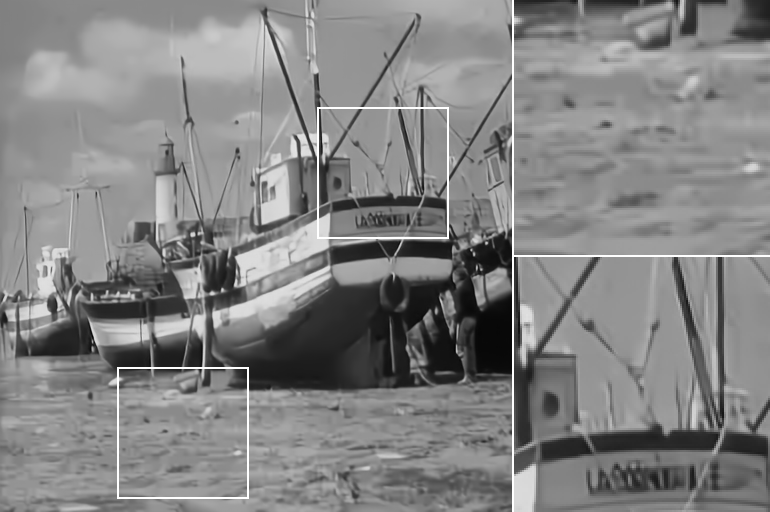}
        \caption{CAS-CNN 29.872/29.656/0.8252}
    \end{subfigure}%
\\
    \begin{subfigure}[b]{\subfigWidth}
		\includegraphics[width=\linewidth]{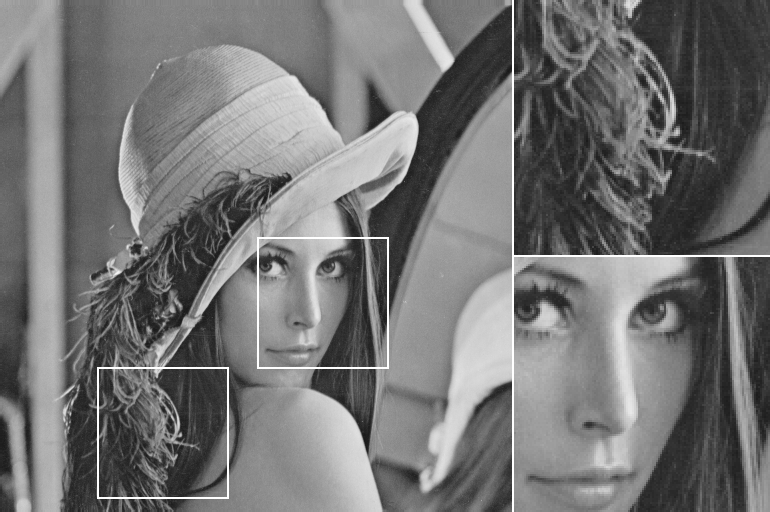}
        \caption{uncompressed}
    \end{subfigure}%
    \hspace*{\fill}
    \begin{subfigure}[b]{\subfigWidth}
		\includegraphics[width=\linewidth]{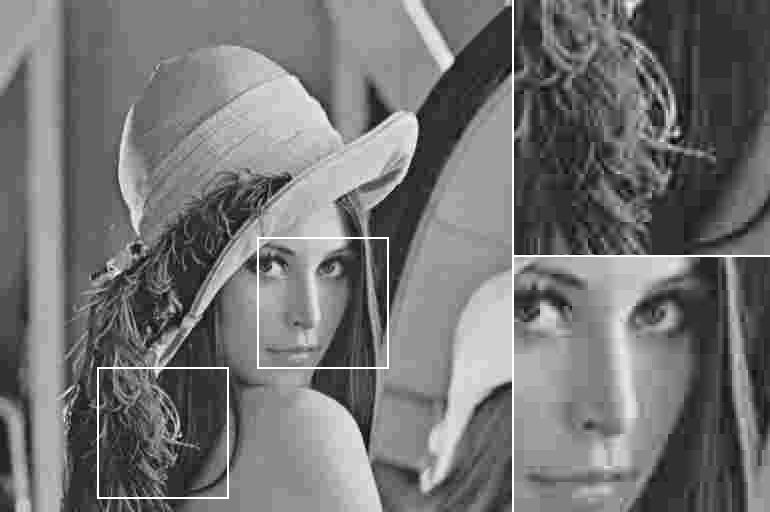}
        \caption{JPEG QF\,10 29.872/29.656/0.8252}
    \end{subfigure}%
    \hspace*{\fill}
    \begin{subfigure}[b]{\subfigWidth}
		\includegraphics[width=\linewidth]{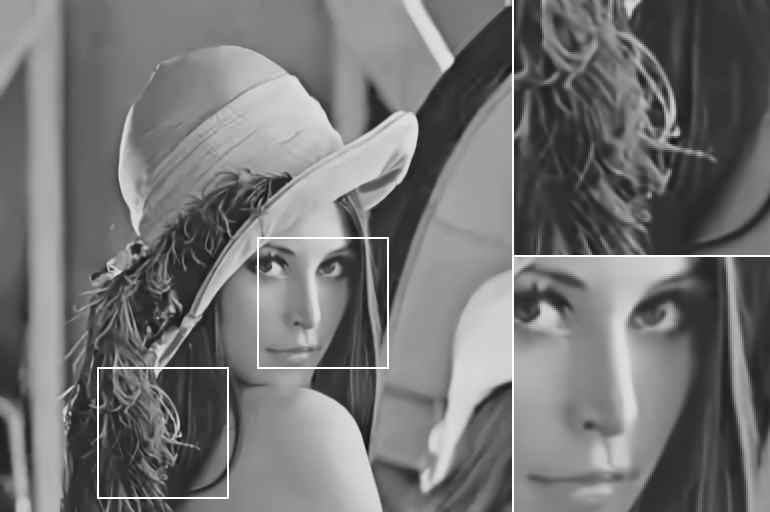}
        \caption{CAS-CNN 32.634/32.414/0.8834}
    \end{subfigure}%
\\
    \begin{subfigure}[b]{\subfigWidth}
		\includegraphics[width=\linewidth]{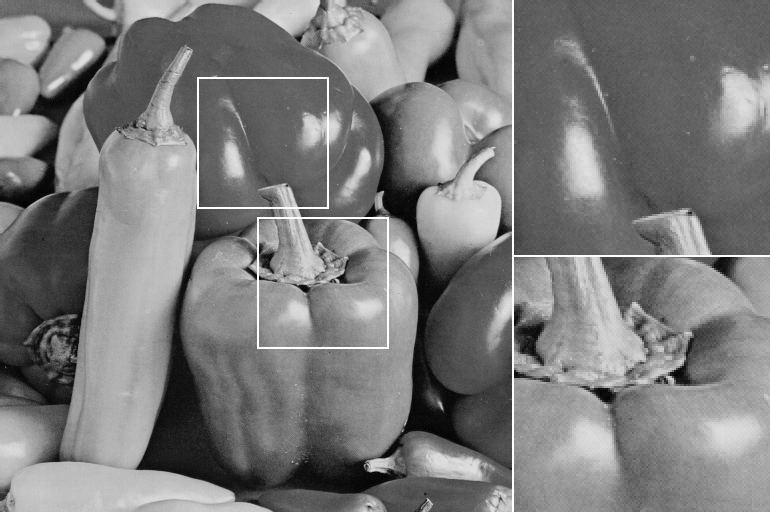}
        \caption{uncompressed}
    \end{subfigure}%
    \hspace*{\fill}
    \begin{subfigure}[b]{\subfigWidth}
		\includegraphics[width=\linewidth]{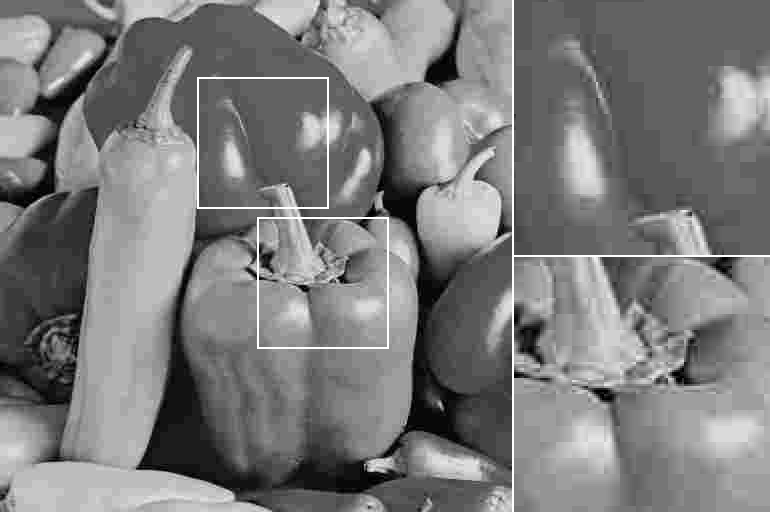}
        \caption{JPEG QF\,10 30.440/27.655/0.8018}
    \end{subfigure}%
    \hspace*{\fill}
    \begin{subfigure}[b]{\subfigWidth}
		\includegraphics[width=\linewidth]{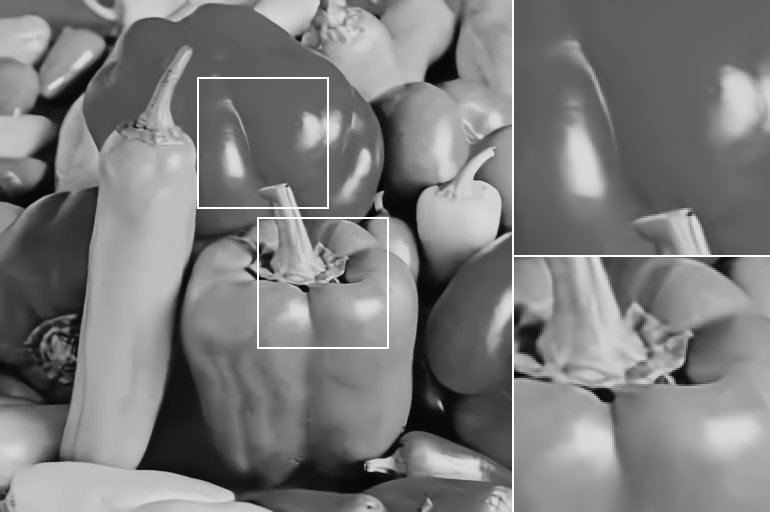}
        \caption{CAS-CNN 32.587/32.437/0.8562}
    \end{subfigure}%
    \caption{Evaluation on the 5 classical test images. We show the uncompressed images (left), the Matlab JPEG QF\,10 compressed images (center), and the result of applying our CAS-CNN to the compressed images. The PSNR/PSNR-B/SSIM with respect to the uncompressed images is indicated below the images.}
    \label{fig:classical5}
\end{figure*}

As discussed in Section~\ref{sec:metrics}, the visual perception can differ from quantitative evaluations using classical quality measures. To give a visual impression as well, we provide a qualitative visual comparison in Figure~\ref{fig:visualComp}. The \emph{lighthouse3} image serves as a basis for this comparison and is the same one used in \cite{Svoboda2016}. It is shown with black markers in Figure~\ref{fig:psnrBppPlot}, indicating that this image is not a particularly well-working outlier. A clear improvement is visible, there are no perceptible blocking artifacts anymore and the ringing artifacts are strongly suppressed without blurring the railing depicted in the image. 
For completeness, we also provide the results for the 5 classical test images used throughout many compression papers (cf. Figure~\ref{fig:classical5}). The trained models and scripts required to reproduce these images are available online\footnote{
\url{http://iis.ee.ethz.ch/~lukasc/cascnn/}}.

In Figure~\ref{fig:multiQfEval}, we show that the networks trained for a specific quality factor do not need to be retrained for the specific quality factor with which the image was compressed to achieve a high improvement in PSNR or PSNR-B. The network trained for QF~60 already boosts the PSNR by more than 1.5\,dB for quality factors ranging from 25 to almost 60. This resilience to variations in quantization has not been shown for approaches focusing on DCT-domain recovery.

\section{Conclusion}
We have presented a 12-layer deep convolutional neural network for compression artifact suppression in JPEG images with hierarchical skip connections and trained with a multi-scale loss function. The result is a new state-of-the-art ConvNet achieving a boost of up to 1.79\,dB in PSNR over ordinary JPEG and showing an improvement of up to 0.36\,dB over the best previous ConvNet result. 
We have shown that a network trained for a specific quality factor is resilient to the QF used compress the input image---a single network trained for QF 60 provides a PSNR gain of more than 1.5\,dB over the wide QF range from 40 to 76. The obtained results are also qualitatively superior to those of existing ConvNets. The network is not tailored to the JPEG-specific compression procedure, and can thus potentially be applied to a wide range of image compression algorithms.

\section*{Acknowledgments}
The authors would like to thank Thilo Weber and Jonas Wiesendanger for their preliminary explorations on this topic, and \emph{armasuisse Science \& Technology} for funding this research.

\bibliographystyle{IEEEtran}
\bibliography{IEEEabrv,mendeley}

\end{document}